\documentclass[runningheads]{llncs}
\usepackage[T1]{fontenc}
\usepackage{booktabs}
\usepackage{multirow}
\usepackage{amsmath}
\usepackage{amssymb}
\usepackage{graphicx,verbatim}

\begin{document}
\title{EchoVQA: Enabling Conversational Assistance for Point-of-Care Cardiac Ultrasound}
%

\author{
Filippos Bellos\inst{1}\and
Yutong Li\inst{1}\thanks{Contributed equally.}\and
Jessie N Dong\inst{2}\textsuperscript{\fnsymbol{footnote}}\and
Zaiyang Guo\inst{2}\textsuperscript{\fnsymbol{footnote}}\and
Emily Mackay\inst{2}\and
Yayuan Li\inst{1}\and
Yannis Avrithis\inst{3}\and
Alison Pouch\inst{2}\and
Jason J. Corso\inst{1}
}

\authorrunning{F. Bellos et al.}

\institute{
University of Michigan\\
\email{\{fbellos,ytongl,yayuanli,jjcorso\}@umich.edu}
\and
University of Pennsylvania\\
\email{\{njdong,aarongzy\}@seas.upenn.edu}\\
\email{\{emily.mackay,pouch\}@pennmedicine.upenn.edu}
\and
Independent Scientist\\
\email{yannis@avrithis.net}
}

\maketitle              
\begin{abstract}
Point-of-care transthoracic echocardiography (TTE) enables cardiac assessment in virtually any clinical setting, yet its diagnostic utility remains constrained by the expertise required for image acquisition and interpretation. Visual question answering (VQA) offers a promising paradigm for bridging this expertise gap through interactive clinical assistance, but existing echocardiography VQA datasets are limited in scale, restricted to high-quality images, and only cover a few views. We introduce EchoVQA, the first large-scale VQA dataset for echocardiography, comprising 14,299 images and 74,819 question-answer pairs. The dataset integrates public sources (EchoNet-Dynamic, CAMUS) with our own point-of-care acquisitions from two handheld probes (Lumify, Clarius), spanning diverse views and including both high-quality and suboptimal images. 
Uniquely, EchoVQA includes acquisition guidance questions to help users optimize transducer positioning toward a diagnostic apical 4-chamber view for left ventricular ejection fraction estimation---a challenging task for novice operators in point-of-care settings. We further develop a parameter-efficient method based on multimodal learnable prompts
achieving state-of-the-art performance on most benchmarks, including EchoVQA, with significantly less trainable parameters than existing state-of-the-art approaches.

\end{abstract}

\section{Introduction}
\label{sec:intro}

Transthoracic echocardiography (TTE) is the first-line imaging modality for assessing cardiac function due to its portability, lack of ionizing radiation, and ease of rapid acquisition~\cite{cheitlin_accaha_1997}. Point-of-care TTE platforms---comprising handheld transducers connected to smartphones or tablets---have expanded cardiac imaging access to resource-limited settings~\cite{kornelsen_rural_2023}. However, interpreting echocardiographic images remains highly operator-dependent, requiring simultaneous identification of standard views, assessment of image quality, evaluation of chamber visibility, and determination of suitability for downstream measurements such as left ventricular ejection fraction (LVEF). LVEF is a key metric of systolic function commonly assessed from the apical 4-chamber (A4C) view, yet optimizing transducer positioning for high-quality A4C acquisition is challenging for novice users~\cite{bick_comparison_2013,mitchell_guidelines_2019}. 
In point-of-care settings, trained sonographers are often unavailable, and clinicians must acquire and interpret images with limited expertise and computational resources.

Visual Question Answering (VQA) has emerged as a powerful paradigm bridging computer vision and natural language processing to provide interactive assistance for medical image analysis~\cite{lin_medical_2023}. 
Several medicine-related benchmark datasets have been introduced: VQA-RAD~\cite{lau2018vqarad} provides 3,515 QA pairs across radiology images, SLAKE~\cite{liu2021slake} offers bilingual physician-verified annotations, PathVQA~\cite{he-etal-2021-towards} contributes 32,799 pairs for pathology, and recent large-scale efforts such as MedTrinity-25M~\cite{xie2025medtrinity} provide millions of image-text pairs across diverse modalities. However, none of these datasets specifically target echocardiography. 
To our knowledge, the only existing echocardiography VQA benchmark~\cite{PhysioNet-mimic-iv-ext-echoqa-1.0.0} provides just 622 QA pairs from high-quality hospital acquisitions, insufficient for training and lacking acquisition guidance for point-of-care settings.

Enabling conversational assistance for medical imaging requires not only comprehensive datasets but also capable models. Multimodal large language models (MLLMs) have emerged as a promising solution. LLaVA-Med~\cite{li2023llava_med} adapts general-domain vision-language models to biomedicine through curriculum learning on figure-caption pairs followed by instruction-tuning. 
LLaVA-Tri~\cite{xie2025medtrinity} further improves performance through pretraining on MedTrinity-25M's multigranular annotations. However, these approaches require training 7–8B parameters, posing significant computational demands. Parameter-efficient fine-tuning (PEFT) methods aim to reduce this burden—PeFoMed~\cite{he2024pefomed} fine-tunes an MLLM with Low-Rank-Adaptation (LoRA) using only 33M trainable parameters---yet underperforms fully fine-tuned approaches.

In this work, we address these challenges with two contributions. First, we introduce EchoVQA, the largest VQA dataset for echocardiography, comprising 14,299 images and 74,819 question-answer pairs. The dataset features multi-turn conversations covering view classification, image quality assessment, chamber visibility, LVEF estimation feasibility, and acquisition guidance for optimizing cardiac views. 
Second, we propose a parameter-efficient method for adapting multimodal large language models to medical VQA, combining visual adaption prompts with text prompt learning~\cite{liu2022ptuningv2}---previously unexplored in multimodal medical settings---and instance-level prompt optimization. Our method achieves state-of-the-art performance on most Med-VQA benchmarks with significantly fewer trainable parameters than LLaVA-Med (7B) and LLaVA-Tri (8B), while substantially outperforming PeFoMed.

\section{EchoVQA Dataset}
\subsection{Data Collection and Pre-Processing.}
We collect echocardiographic images from two complementary sources: public datasets and our own point-of-care acquisitions.
\subsubsection{Public Data.} 
\begin{figure}[!t]
\includegraphics[width=\textwidth,trim=0 164 140 164,clip]{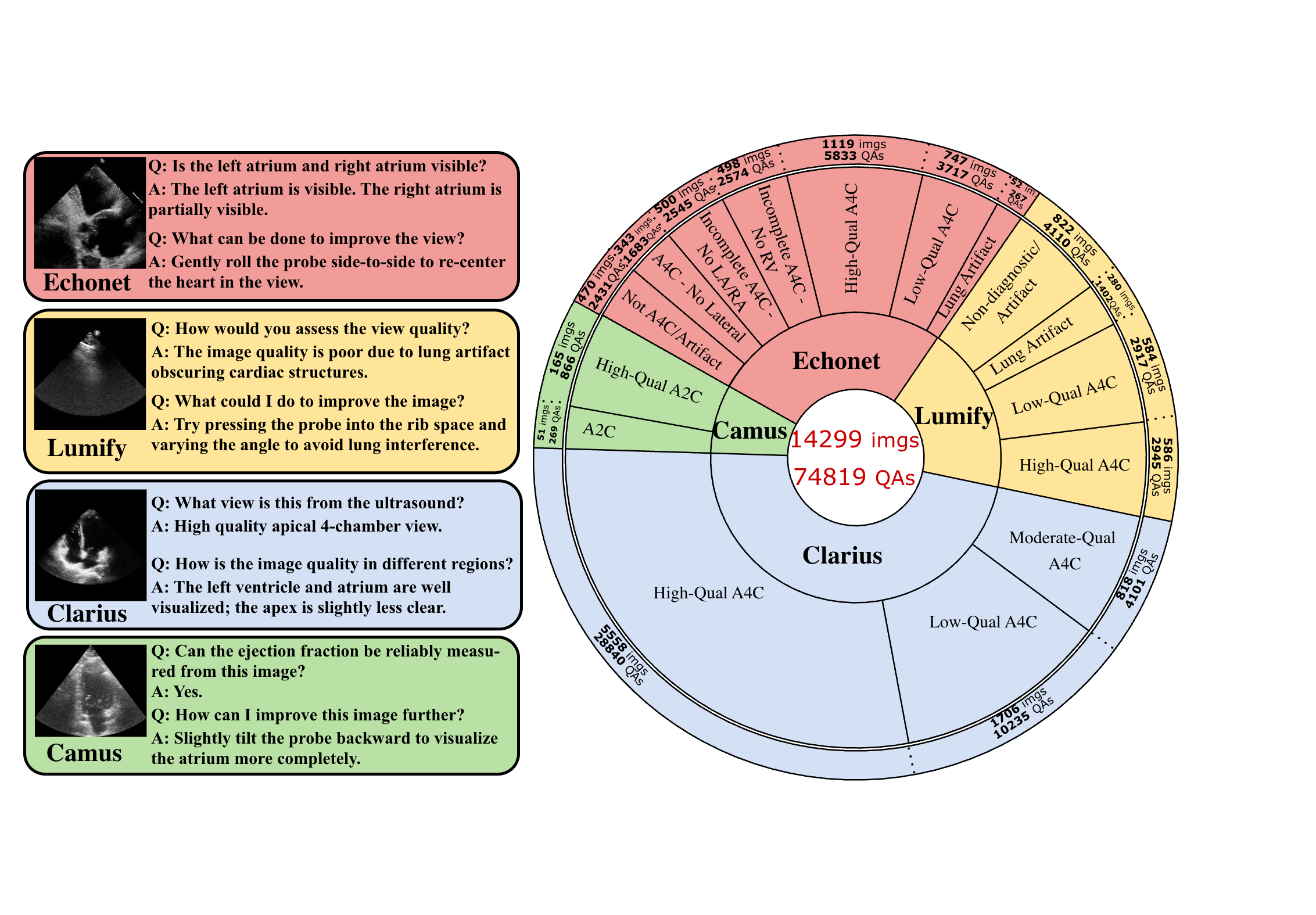}
\caption{EchoVQA dataset overview.  \textbf{Left}: Example QA pairs from each data source demonstrating the different question types. \textbf{Right}: Distribution of images and QA pairs across data sources (inner ring) and quality categories (outer ring). The dataset comprises 14,299 images and 74,819 QA pairs spanning public (EchoNet-Dynamic, CAMUS) and our point-of-care (Lumify, Clarius) acquisitions.} \label{dataset_stats}
\end{figure}

Public datasets such as EchoNet-Dynamic~\cite{ouyang_video-based_2020} and CAMUS~\cite{leclerc2019camus} are acquired by trained sonographers in hospital echo labs using cart-based systems. 
Our inspection reveals substantial variation within EchoNet-Dynamic, including non-standard views (e.g. A4C missing structures), lung artifacts, etc.) (Fig.~\ref{dataset_stats}). We explicitly examine and categorize these variations to enable targeted question-answer generation.

\subsubsection{Point-of-Care Data.} To complement public data with realistic point-of-care acquisitions, typically performed with handheld probes by non-specialist clinicians, we collect additional scans using the Philips Lumify and Clarius PAL HD3, under institutional IRB approval. Using the Lumify, two operators performed scans on 3 subjects, and using the Clarius, scans were collected from 9 subjects. These acquisitions capture the full spectrum of image quality a novice sonographer would encounter, stratified into five categories reflecting the progression toward a diagnostic A4C view for LVEF estimation: (1)\textit{high-quality A4C}---all four chambers clearly visible, LVEF readily measurable; (2)\textit{moderate-quality A4C}---partial chamber visibility with medium image quality, LVEF generally measurable; (3)\textit{low-quality A4C}---recognizable A4C but compromised by suboptimal visualization, missing structures, off-axis orientation, or poor image settings, LVEF typically not feasible; (4)\textit{lung artifact}---rib shadows or pleural interfaces obscuring cardiac structures; and (5)~\textit{non-diagnostic/artifacts}---inadequate for clinical interpretation due to severe artifacts or cardiac structures not in view. The specific nuances within each category are captured in the corresponding QA pairs.

\begin{figure}[!t]
\includegraphics[width=\textwidth,trim=0 10 0 0,clip]{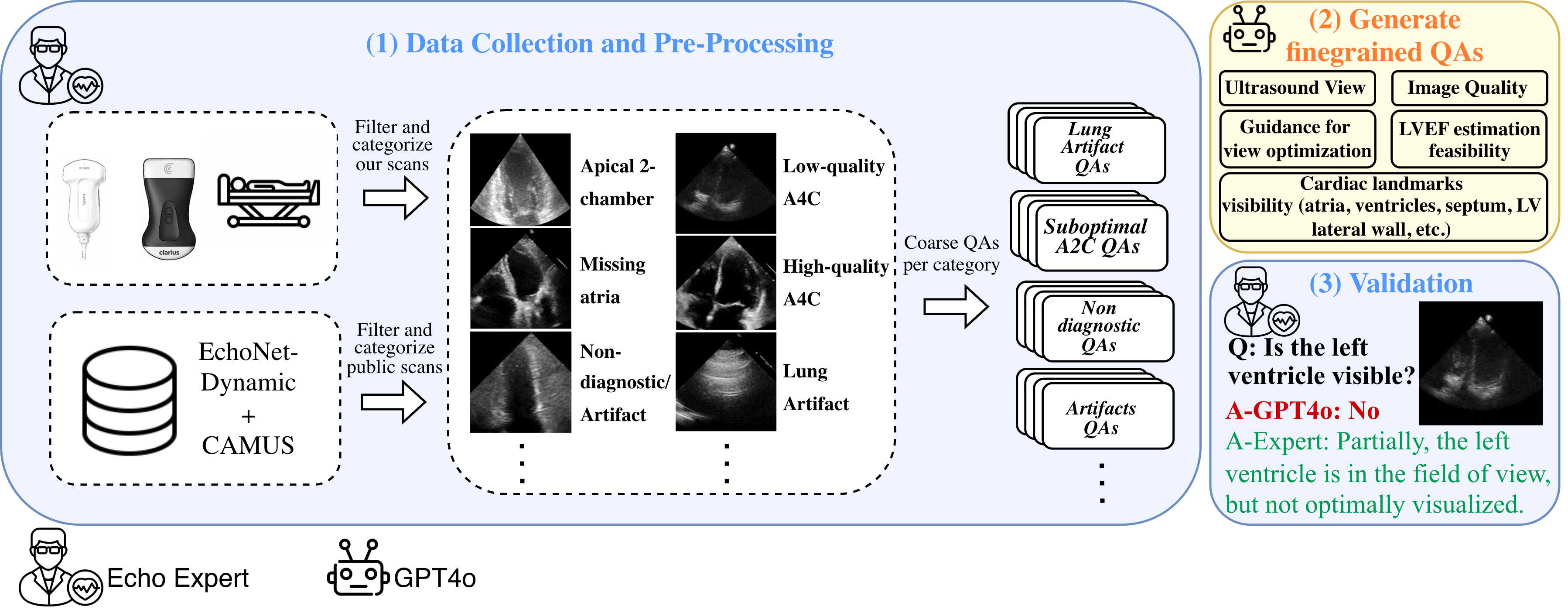}
\caption{EchoVQA construction pipeline. (1) Echocardiography experts categorize images from our point-of-care acquisitions and public datasets into view and quality-based categories, then define coarse QA templates for each category. (2) GPT-4o generates diverse, image-specific QA pairs conditioned on category labels and expert templates, covering cardiac landmark visibility, LVEF estimation feasibility, ultrasound view classification, acquisition guidance, and image quality. (3) Experts review all generated QAs, correcting inaccurate or poorly grounded responses to ensure clinical accuracy.} \label{dataset_pipeline}
\end{figure}

\subsection{Question-Answer Generation.}
We develop a human-in-the-loop annotation pipeline that ensures clinical accuracy while enabling efficient scaling through LLM assistance (Figure~\ref{dataset_pipeline}).

\textbf{Category-Level Expert Templates.} For each image category, echocardiography experts define a set of template question-answer pairs that establish the clinical scope and ground truth for that category. These templates encode expert knowledge about quality assessment, expected chamber visibility, LVEF estimation feasibility, and acquisition guidance—ranging from correcting suboptimal views toward a diagnostic A4C (e.g., ``Adjust the probe position laterally and angle it toward the patient's left shoulder'') to further optimizing already diagnostic images (e.g., ``Rotate the probe slightly clockwise to better visualize the septum and right atrium''). These expert-authored templates serve as conditioning context for scalable QA generation.

\textbf{Fine-Grained QA Generation.} Using GPT-4o, we generate diverse, image-specific question-answer pairs conditioned on the category label and its associated expert templates. Critically, while templates provide clinical grounding, the model is prompted to analyze each image directly, as substantial variability exists even within the same category. For instance, images categorized as low-quality A4C may exhibit, among other issues, low contrast obscuring chamber boundaries, partial left ventricular cutoff compromising LVEF estimation while other chambers remain visible, or foreshortening that distorts the true apex position. The model generates questions spanning five types for each ultrasound image: view classification, chamber visibility assessment, image quality evaluation, LVEF estimation feasibility, and acquisition guidance for view optimization. This approach enables efficient scaling---generating thousands of QA pairs per category---while ensuring all generated content remains anchored to both expert-validated clinical knowledge and image-specific observations. 
\subsection{Expert Validation.}Although MLLMs have shown promise in medical imaging, they remain imperfect at specialized medical tasks and may introduce systematic biases in generated content. It is therefore imperative that all generated QA pairs undergo expert review. Four echocardiography experts reviewed the full set of generated QA pairs. For each pair, reviewers verified factual correctness, ensured answers were grounded in the specific image rather than reflecting generic category-level knowledge, and corrected or rewrote responses that failed to meet these criteria. They further refined answer phrasing for clinical precision and rejected questions that were ambiguous or inappropriate for a given image.  
This validation step ensures that despite LLM-assisted generation, the final dataset meets the quality standards expected for clinical use.

\section{Parameter-Efficient Medical VQA}
\label{sec:method}
Current medical VQA methods operate at a single level of adaptation---modifying either full model weights or low-rank subspaces uniformly. We instead propose a parameter-efficient multi-level prompt framework that injects task-relevant information at three complementary levels: (1) visual adaption prompts~\cite{zhang_llama-adapter_2023} that inject visual information into the language model, (2) dataset-level text prompts that encode shared domain knowledge via prompt learning~\cite{liu2022ptuningv2}---previously unexplored in multimodal medical settings, and (3) instance-level prompts optimized at inference to capture image-specific context.
\begin{figure}[!t]
\includegraphics[width=\textwidth,trim=0 30 0 30,clip]{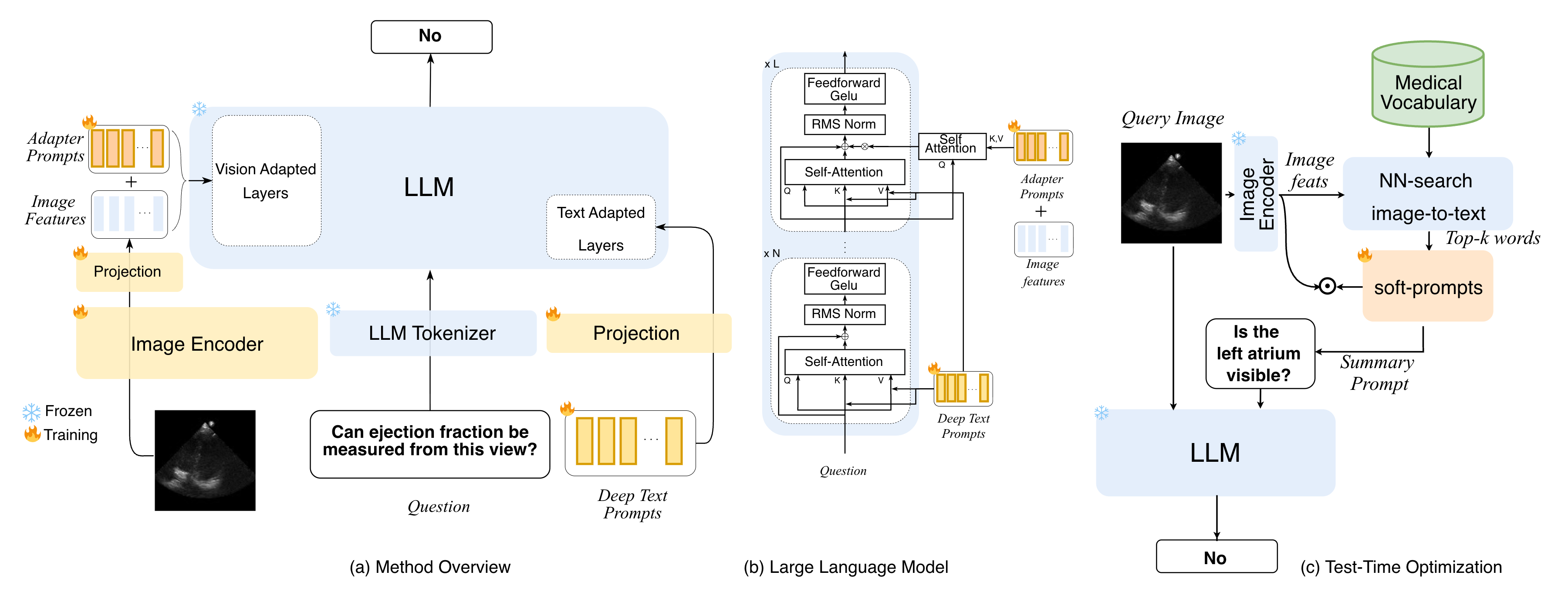}
\caption{Proposed method. \textbf{(a)} Training pipeline with visual adaption prompts injected into top LLM layers. \textbf{(b)} Dataset-level text prompts added to keys and values in the layer's self-attention. \textbf{(c)} At inference, instance-level prompts are retrieved from a medical vocabulary, optimized, and injected as a summary prompt.}\label{method}
\end{figure}
\subsection{Model Architecture}
The proposed method and model architecture is illustrated in Figure~\ref{method}.

\textbf{Vision Encoder.}
The input echocardiography image $X^v$ is encoded by a vision encoder $f^v$ to obtain visual features $Z^v := f^v(X^v) \in \mathbb{R}^{D \times N_v}$, where $D$ is the embedding dimension and $N_v$ is the number of visual tokens. These features are projected to the LLM embedding space via a learnable projection $W_p \in \mathbb{R}^{D_L \times D}$, yielding $\tilde{Z}^v := W_p Z^v \in \mathbb{R}^{D_L \times N_v}$.

\textbf{Text Embedding.}
The input question $X^q$ is tokenized into a sequence of $S$ tokens and mapped by the frozen LLM embedding layer $f^t$ to obtain text embeddings $Z^t := f^t(X^q) = (z^t_1, \ldots, z^t_S) \in \mathbb{R}^{D_L \times S}$.

\textbf{Visual Adaption Prompts.} Following~\cite{zhang_llama-adapter_2023}, we inject visual information into the LLM via learnable adaption prompts $P^a \in \mathbb{R}^{D_L \times N_a}$. The projected visual features are pooled into a global token and added to $P^a$, then prepended to input sequences at the top $L'$ transformer layers. We use zero-initialized gating with learnable scalars $\alpha^{(l)}$ for stable training.

\textbf{Dataset-Level Text Prompts.}
Beyond visual adaption prompts, we introduce learnable text prompts that encode domain-specific knowledge shared across the dataset. These prompts are prepended to the keys and values in the self-attention mechanism. Given an input sequence $H \in \mathbb{R}^{D_L \times K}$, the query, key, and value are computed as:
$Q := W^Q H, \quad K := [P^K; W^K H], \quad V := [P^V; W^V H],$
where $W^Q, W^K, W^V \in \mathbb{R}^{D_L \times D_L}$ are the projection matrices, and $P^K, P^V \in \mathbb{R}^{D_L \times N_p}$ are learnable text prompt parameters with $N_p$ prompt tokens. This formulation allows the model to attend to learned domain-specific context without modifying the output sequence length.

\subsection{Training}
We adopt a two-stage training strategy following prior work in medical multimodal learning~\cite{he2024pefomed,li2023llava_med,xie2025medtrinity}. In \textbf{Stage 1}, we train on medical image-caption datasets to align visual representations with the language model's embedding space. In \textbf{Stage 2}, we fine-tune on downstream medical VQA datasets. The model is trained with the standard language modeling objective:
\begin{equation}
\mathcal{L} = -\sum_{i=1}^{|X^a|} \log p(x^a_i | X^v, X^q, x^a_{<i}; \Theta),
\end{equation}
where $\Theta$ denotes all trainable parameters and $x^a_{<i}$ represents answer tokens preceding position $i$.
During both stages, we update the vision encoder $f^v$, visual projection $W_p$, adaption prompts $P^a$, gating scalars $\{\alpha^{(l)}\}$, and text prompt key-value pairs $\mathcal{P}^t$, while keeping the LLM frozen.

\subsection{Inference: Instance-Level Prompt Optimization}

We introduce instance-level test time prompt optimization to inject image-specific context beyond the dataset-level prompts learned during training.

We construct a \textbf{medical vocabulary} $\mathcal{V} = \{v_1, v_2, \ldots, v_M\}$ containing $M$ domain-specific terms from the train set of each downstream dataset. Each term $v_m$ is encoded using BiomedCLIP's text encoder to obtain vocabulary embeddings $E^{\mathcal{V}} \in \mathbb{R}^{D \times M}$. Given a query image $X^v$, we retrieve the top-$k$ most similar vocabulary terms via cosine similarity:
$\mathcal{R} = \text{top-}k \left( \frac{Z^v \cdot E^{\mathcal{V}}}{\|Z^v\| \|E^{\mathcal{V}}\|} \right),$
where $\mathcal{R} = \{r_1, \ldots, r_k\}$ denotes the retrieved terms.

The retrieved terms $\mathcal{R}$ are tokenized and mapped to the LLM embedding space via the frozen embedding layer to obtain $P^{init} := f^t(\mathcal{R}) \in \mathbb{R}^{D_L \times N_r}$, where $N_r$ is the total number of tokens. These embeddings initialize soft prompts $P^{soft}$, which are \textbf{optimized for $T$ iterations} to better align with the visual features:
$P^{soft*} = \arg\min_{P^{soft}} \mathcal{L}_{align}(P^{soft}, Z^v),$
where $\mathcal{L}_{align}$ is a cosine similarity loss. After optimization, we select the first retrieved term's embedding as a compact summary prompt capturing the dominant visual semantics. This single summary prompt is concatenated with the question embeddings
$\hat{Z}^t = [P^{summary}; Z^t]$,
and the final answer is generated by the LLM conditioned on visual features $\tilde{Z}^v$ and the augmented text embeddings $\hat{Z}^t$.

\section{Experiments}
\label{sec:experiments}
We conduct experiments to evaluate the effectiveness of our proposed method on established medical VQA benchmarks and our EchoVQA dataset.

\textbf{Implementation Details.}
We use BiomedCLIP~\cite{zhang2023biomedclip} as our vision encoder and LLaMA-2-7B~\cite{touvron_llama_2023} as the language model. The LLM remains frozen during training while the vision encoder is fine-tuned. We set the number of visual adaption prompts $N_a = 10$ and text prompt tokens $N_p = 10$. 

In Stage 1, we train on medical image-caption datasets (ROCO~\cite{roco}, ImageCLEF~\cite{clef2022}, MEDICAT~\cite{subramanian-etal-2020-medicat}, and MIMIC-CXR~\cite{mimic_cxr}). In Stage 2, we fine-tune on each downstream VQA dataset following~\cite{li2023llava_med,xie2025medtrinity,he2024pefomed}.
For test-time prompt optimization, we retrieve $k=10$ terms from the medical vocabulary and optimize for $T=100$ iterations before selecting the top term as the summary prompt. All experiments are conducted on 2 NVIDIA H100 GPUs.

\textbf{Evaluation.}
We evaluate on three established medical VQA benchmarks: VQA-RAD~\cite{lau2018vqarad}, SLAKE~\cite{liu2021slake}, and PathVQA~\cite{he-etal-2021-towards}. We additionally evaluate on our proposed EchoVQA benchmark. Following ~\cite{li2023llava_med,xie2025medtrinity}, we report recall for open-ended questions and exact match accuracy for closed-ended questions.

We compare against several state-of-the-art \textbf{baselines}: GPT-4V~\cite{openai2023gpt4} as a zero-shot baseline; LLaVA~\cite{liu2023llava} as a general-domain MLLM; LoRA finetuning (PeFoMed)~\cite{hu2022lora,he2024pefomed}, LLaVA-Med~\cite{li2023llava_med} and LLaVA-Tri~\cite{xie2025medtrinity}.

\begin{table}[t]
\centering
\caption{Performance comparison on medical VQA benchmarks. We report recall for open-ended questions and exact match accuracy for closed-ended questions. All methods except GPT-4V have been reproduced by us, using their respective codebases.}
\label{tab:results}
\footnotesize
\begin{tabular}{lc*{8}{c}}
\toprule
\multirow{2}{*}{Method} & \multirow{2}{*}{Params} & \multicolumn{2}{c}{VQA-RAD} & \multicolumn{2}{c}{SLAKE} & \multicolumn{2}{c}{PathVQA} & \multicolumn{2}{c}{EchoVQA} \\
\cmidrule(lr){3-4} \cmidrule(lr){5-6} \cmidrule(lr){7-8} \cmidrule(lr){9-10}
 & & Open & Closed & Open & Closed & Open & Closed & Open & Closed \\
\midrule
GPT-4V & -- & 39.5 & 78.9 & 33.6 & 43.6 & -- & -- & -- & -- \\
LLaVA & 7B & 63.7 & 83.4 & 83.7 & 84.6 & 37.1 & 90.8 & 56.0 & 70.6 \\
PeFoMed (LoRA) & 33M & 58.5 & 83.4 & 81.3 & 82.7 & 29.5 & 88.4 & 54.4 & 86.9 \\
LLaVA-Med & 7B & 64.7 & 84.2 & 83.0 & 85.3 & \textbf{38.5} & 91.2 & 55.5 & 82.2 \\
LLaVA-Tri & 8B & 65.1 & 84.9 & 82.3 & 83.2 & 37.8 & \textbf{91.8} & 55.2 & 79.7 \\
\midrule
Ours & 104M & \textbf{65.8} & \textbf{86.8} & \textbf{87.3} & \textbf{89.7} & 37.2 & 91.3 & \textbf{60.2} & \textbf{90.4} \\
\bottomrule
\end{tabular}
\end{table}

\textbf{Results.}
Table~\ref{tab:results} presents results on established medical VQA benchmarks. Our method achieves state-of-the-art performance on VQA-RAD, SLAKE and EchoVQA, outperforming all baselines.
On PathVQA, we achieve competitive but not state-of-the-art performance. We attribute this to the domain gap: PathVQA comprises diverse pathology images, while Stage 1 pretraining datasets are predominantly radiology-focused. Methods that fully fine-tune the LLM can better adapt to pathology-specific patterns, whereas our frozen LLM approach might rely more on the alignment learned during pretraining. 

\begin{table}
\centering
\caption{Ablation study evaluating the contribution of visual adaption prompts (AP), dataset-level text prompts (TP), and test-time optimization (TTO).}\label{tab:ablation}
\footnotesize
\begin{tabular}{lcccccccc}
\toprule
\multirow{2}{*}{Method} & \multicolumn{2}{c}{VQA-RAD} & \multicolumn{2}{c}{SLAKE} & \multicolumn{2}{c}{PathVQA} & \multicolumn{2}{c}{EchoVQA} \\
\cmidrule(lr){2-3} \cmidrule(lr){4-5} \cmidrule(lr){6-7} \cmidrule(lr){8-9}
 & Open & Closed & Open & Closed & Open & Closed & Open & Closed \\
\midrule
AP & 65.6 & 80.6 & 87.1 & 88.1 & 36.9 & 90.3 & 60.0 & 88.7 \\
AP + TP & 65.6 & 85.7 & \textbf{87.3} & 89.4 & \textbf{37.2} & 91.2 & 60.0 & \textbf{90.4} \\
AP + TP + TTO (Full) & \textbf{65.8} & \textbf{86.8} & \textbf{87.3} & \textbf{89.7} & \textbf{37.2} & \textbf{91.3} & \textbf{60.2} & \textbf{90.4} \\
\bottomrule
\end{tabular}
\end{table}

Table~\ref{tab:ablation} isolates the contribution of each component. Visual adaption prompts alone provide a strong baseline. Adding dataset-level text prompts yields consistent gains on closed-ended questions across all benchmarks, with the largest improvement on VQA-RAD (+5.1 closed). Test-time optimization provides further incremental gains.

\section{Conclusion}
We introduced EchoVQA, the first large-scale VQA dataset for echocardiography, featuring 14,299 images and 74,819 QA pairs spanning diverse views, quality levels, and acquisition guidance for point-of-care settings. We further proposed a parameter-efficient method combining visual adaption prompts, text prompt learning, and test-time optimization that achieves state-of-the-art performance not only on EchoVQA but on other standard medical VQA benchmarks.

\begin{credits}
\subsubsection{\ackname}
This research was funded, in part, by the U.S. Government, under Agreement No. 1AY2AX000062. The views and conclusions contained in this document are those of the authors and should not be interpreted as representing the official policies, either expressed or implied, of the U.S. Government.
\end{credits}

\bibliographystyle{splncs04}
\bibliography{mybibliography}

\end{document}